\documentclass[10pt,twocolumn,letterpaper]{article}
\usepackage[accsupp]{axessibility} 
\usepackage{wacv}
\usepackage{times}
\usepackage{epsfig}
\usepackage{graphicx}
\usepackage{amsmath}
\usepackage{amssymb}
\usepackage{booktabs}

\usepackage{array}
\usepackage{amsfonts}
\usepackage{amsmath}
\usepackage[switch]{lineno}
\usepackage{multirow}
\usepackage{pifont}
\usepackage{verbatim}

\usepackage[cmyk]{xcolor}
\usepackage{soul}

\def\wacvPaperID{170} 

\wacvfinalcopy 

\ifwacvfinal
\usepackage[breaklinks=true,bookmarks=false]{hyperref}
\else
\usepackage[pagebackref=true,breaklinks=true,colorlinks,bookmarks=false]{hyperref}
\fi

\begin{document}

\title{Cross-modal Semantic Enhanced Interaction for Image-Sentence Retrieval}
\author{Xuri Ge$^1$, Fuhai Chen$^{2}$\thanks{Corresponding author}, Songpei Xu$^1$, Fuxiang Tao$^1$, Joemon M. Jose${^{1}}$\\
$^1$School of Computing Science, University of Glasgow, Glasgow, UK. \\
$^2$Department of Computer Science, The University of Hong Kong, Hong Kong, China.\\
{\tt\small x.ge.2@research.gla.ac.uk, chenfuhai3c@163.com, s.xu.1@research.gla.ac.uk, }\\ 
{\tt\small f.tao.1@research.gla.ac.uk, Joemon.Jose@glasgow.ac.uk}
}

\maketitle
\thispagestyle{empty}

\begin{abstract}
    Image-sentence retrieval has attracted extensive research attention in multimedia and computer vision due to its promising application. 
    The key issue lies in jointly learning the visual and textual representation to accurately estimate their similarity. 
    To this end, the mainstream schema adopts an object-word based attention to calculate their relevance scores and refine their interactive representations with the attention features, which, however, neglects the context of the object representation on the inter-object relationship that matches the predicates in sentences.
    In this paper, we propose a Cross-modal Semantic Enhanced Interaction method, termed \textbf{CMSEI} for image-sentence retrieval, which correlates the intra- and inter-modal semantics between objects and words. 
    In particular, we first design the intra-modal spatial and semantic graphs based reasoning to enhance the semantic representations of objects guided by the explicit relationships of the objects' spatial positions and their scene graph. 
    Then the visual and textual semantic representations are refined jointly via the inter-modal interactive attention and the cross-modal alignment. 
    To correlate the context of objects with the textual context, we further refine the visual semantic representation via the cross-level object-sentence and word-image based interactive attention. 
    Experimental results on seven standard evaluation metrics show that the proposed CMSEI outperforms the state-of-the-art and the alternative approaches on MS-COCO and Flickr30K benchmarks. 
\end{abstract}

\section{Introduction}

    \begin{figure}[t] 
    	\centering
    	\vspace{-1em}
    	\includegraphics[width=1.0\linewidth]{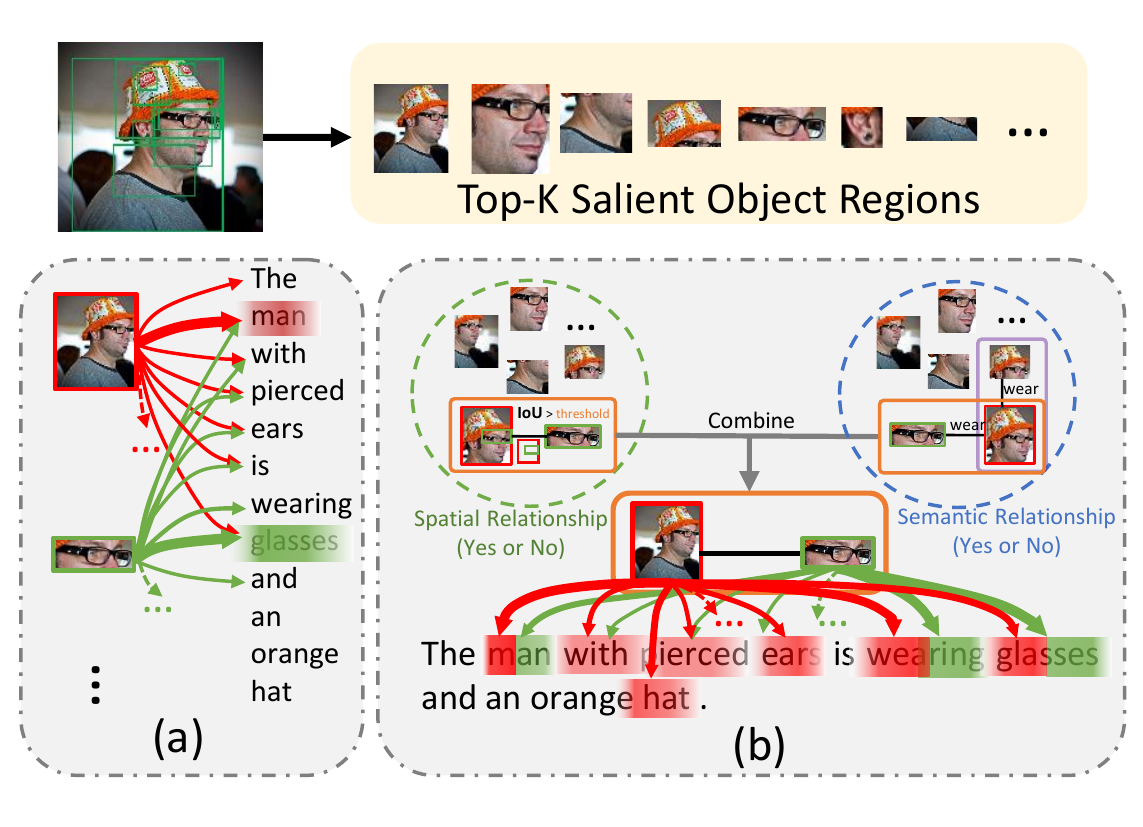}
    	\vspace{-2em}
    	\caption{
    	Illustration of two cross-modal semantic interaction schemas (only show image-to-sentence retrieval for clarity). (a) traditional schema: the correlated objects (\emph{e.g.} \emph{man} and \emph{glass}) are hardly attended to their common word (\emph{e.g.} \emph{wear}) with high relevance score (thick arrow), 
    	(b) our CMSEI method: the relationships between the correlated objects are integrated into the region features of these objects, whose common word is with high relevance score. Note that the semantic relationships are detected via scene graph and judged by whether there is a predicted label (\emph{e.g.} \emph{wear}) with high confidence.
        }
    	\label{fig:frams}
    	\vspace{-1.5em}
    \end{figure}

    Image-sentence retrieval aims at retrieving the most relevant images (or sentences) given a query sentence (or image),  which involves the cross-over study on computer vision and neural language processing \cite{frome2013devise, wang2016learning, faghri2017vse++, lee2018stacked,chen2019variational}.
    Due to its broad applications, such as multimedia analysis, multimedia search, album management, and medical image retrieval, image-sentence retrieval has aroused the widespread research attention. 
    The key issue of image-sentence retrieval lies in jointly learning the visual and textual representations to guarantee their similarity between the matched image and sentence.

    To this end, existing works mainly adopt two schemas to learn the visual and textual representations, \emph{i.e.} modality-independent representation learning \cite{frome2013devise, karpathy2014deep, wang2016learning, faghri2017vse++, zhen2019deep, wen2020learning, ge2021structured, VSEwuqiong,cheng2022cross} and the cross-modal semantic interaction \cite{lee2018stacked, huang2018learning, liu2019focus,DIME,NAAF,Gradual}
    Specially, on one hand, the modal-independent representation learning has been widely studied due to its high retrieval efficiency. 
    For instance, \cite{frome2013devise, wang2016learning, faghri2017vse++, zhen2019deep} optimized a joint embedding space by minimizing the distance of the visual and textual global features, which are directly extracted from the whole image and the full sentence via Convolutional Neural Networks (CNNs) and Recurrent Neural Networks (RNNs) respectively.
    Several recent works \cite{huang2018learning, VSEwuqiong, cheng2022cross} extracted the visual and textual local features from object regions and words and integrated them as a whole respectively before projecting these two features into a common latent space. 
    However, due to the lack of the deep semantic correlation on the fine-grained fragments, these methods are limited on the retrieval accuracy. 

    On the other hand, cross-modal semantic interaction is proposed to boost the retrieval performance by learning the accurate visual-textual semantic relevance between the fragments of image and sentence \cite{lee2018stacked, huang2018learning, liu2019focus, DIME,NAAF}, as shown in Figure \ref{fig:frams} (a).
    For instance, SCAN \cite{lee2018stacked} attended object regions to each word to generate the text-aware visual features for sentence-to-image matching and conversely for image-to-sentence matching. 
    Although achieving the significant improvement, these methods neglects the fact that little inter-object relationships are reflected in the object representations compared to the strong context of the textual structure, which leads to a feeble role of visual semantic during image-sentence matching. 

    To deal with the above problem, it's intuitive to put forward two straightforward solutions to cooperate visual semantic representation with the inter-object relationship. 
    On one hand, the object region features can be concatenated with the feature of the inter-object relationship detected by an off-the-shelf detector \cite{yao2017boosting,wu2017image,wang2020consensus,anderson2018bottom}. However, such method has three defects that affect the retrieval performance: (i) it's hard to keep a structured correlation among the objects and their relationships in a multi-layer network without continuous correlation guidance; (ii) the detected relationship labels, trained on different dataset, bring about the extra recognition error; and (iii) it's not an end-to-end framework. 
    On the other hand, the inter-object relationships can be utilized for object representation enhancement via the graph-based modeling. The representative solutions are in two folds. First, following \cite{wang2020cross,nguyen2021deep}, the relationships are detected guided by the scene graph and their label-based features are aggregated with the object region features to feed the Graph Convolution Networks (GCNs). However, such methods still suffer from the aforementioned (ii) and (iii) as revealed in \cite{Gradual}. Second, following \cite{SGRAF,cheng2022cross}, the relationships are implicitly reflected via the fully-connected GCNs where the object region features are input as graph nodes, nevertheless, leaving the relationship information weak and ambiguous that effects the object discrimination. Therefore, it's natural for us to consider an integrated structured modeling that captures the explicit information of the inter-object relationships to enhance the object representation. As manifested in Figure \ref{fig:frams} (b), by explicitly constructing the inter-object relationship, it's easier compared to \ref{fig:frams} (a) for the correlated object (\emph{e.g.} regions of \emph{head} and \emph{glasses}) to obtain the high relevance with the correlated predicate words (\emph{e.g.} word \emph{wear}). 
    
    
    Driven by the above consideration, we propose a novel cross-modal semantic enhanced interaction method for image-sentence retrieval, termed \textit{CMSEI}, which correlates the intra- and inter-modal semantics between objects and words.
    For the intra-modal semantic correlation, the inter-object relationships are explicitly reflected on the spatially relative positions and the scene graph guided potential semantic relationships among the object regions. We then propose a relationship-aware GCNs model (termed \textit{R-GCNs}) to enhance the object region representations with their relationships, where the graph nodes are object region features and the graph structures are determined by the inter-object relationships, \emph{i.e.} each edge connection in the graph adjacency matrices rely on whether there is a relationship with high confidence. Different from \cite{wang2020cross,nguyen2021deep}, intra-modal semantic correlation in CMSEI minimizes the error interference from the detection and maximizes the feasibility of the end-to-end representation learning. 
    For the inter-modal semantic correlation, the semantic enhanced representations of words that undergo a fully-connected GCNs model, as well as the semantic enhanced representations of object regions are attended alternatively in the inter-modal interactive attention, where the object region features are attended to each word to refine its feature and conversely the word feature are attended to each object region to refine its feature. To correlate the context of objects with textual context, we further refine the representations of object regions and words via cross-level object-sentence and word-image based interactive attention. 
    The intra-modal semantic correlation, inter-modal semantic correlation, and the similarity-based cross-modal alignment are jointly executed to further enhance the cross-modal semantic interaction.

    The contributions of this paper are as follows: 
        (1) We explore an intra-modal semantic enhanced correlation to explicitly utilize the inter-object spatially relative positions and inter-object semantic relationships guided by scene graph, and propose a relationship-aware GCNs model (R-GCNs) to enhance the object region features with their relationships. This module mitigates the error interference from the detection and enables the end-to-end representation learning. 
        (2) We propose a cross-modal semantic enhanced interaction method (CMSEI) to unite the intra-modal semantic correlation, inter-modal semantic correlation, and the similarity-based cross-modal alignment to simultaneously model the semantic correlations on three grain levels, \emph{i.e.} intra-fragment, inter-fragment, inter-instance. Especially, cross-level interactive attention is proposed to model the correlations between the fragments and the instance. 
        (3) The proposed CMSEI is sufficiently evaluated with extensive experiments on MS-COCO and Flickr30K benchmarks. The results in seven standard evaluation metrics demonstrate the superiority of the proposed CMSEI, where CMSEI achieves the state-of-the-art on the most of metrics.

    \begin{figure*}[t] 
    	\centering
    	\includegraphics[width=1\linewidth]{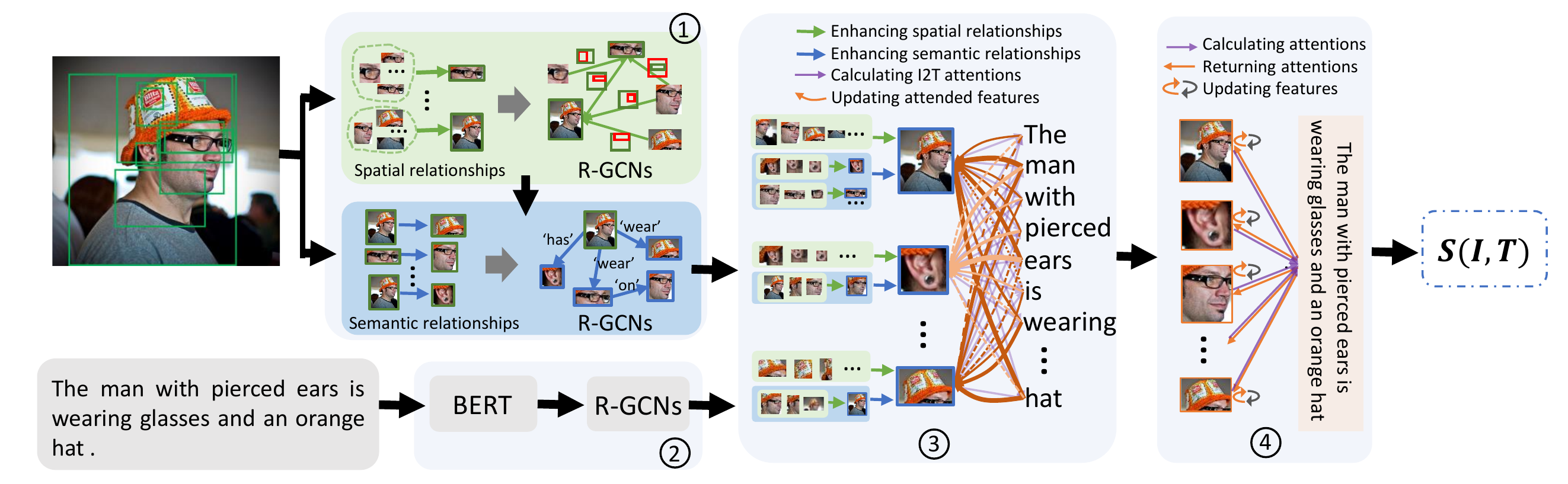}
    	\vspace{-2em}
    	\caption{
    	The overall framework (image-to-sentence version) of CMSEI. In intra-modal semantic correlation (\ding{172}), two relationship-aware GCNs are constructed to respectively integrate the explicit spatial and semantic relationships between each two objects into their region representations by changing the relationship-determined graph adjacency matrices. In \ding{173}, a pre-trained BERT model is used to obtain high-level semantic embedding features of words, which are then fed to GCNs to enhance them with the context. In inter-modal semantic correlation (\ding{174} and \ding{175}), the visual and textual semantic features are further enhanced via object-word interactive attention and the visual semantic representation is refined via the cross-level object-sentence and word-image based interactive attention. Visual and textual semantic similarity is finally estimated for the cross-modal alignment.}
    	\label{fig:fig_overall}
    	\vspace{-1em}
    \end{figure*}

\section{Related Work}
    The key issue of the image-sentence retrieval is measuring the visual-textual similarity between an image and a sentence. 
    It can be divided into two main kinds: modality-independent representation retrieval and cross-modal interaction retrieval. CMSEI belongs to the latter one.
    
    \noindent \textbf{Modality-independent representation retrieval.} 
    Most earlier works \cite{frome2013devise, kiros2014unifying, mao2014deep, fang2015captions, vendrov2015order, wang2016learning, wang2018learning} used independent processing of images and sentences within two branches to obtain a holistic representation of images and sentences. 
    Some works \cite{frome2013devise, kiros2014unifying, wang2016learning, zheng2020dual} directly extracted the features of two modalities from the whole image via CNNs and from the full sentence via RNNs. 
    Inspired by the detection of object regions, many studies \cite{karpathy2014deep, karpathy2015deep} started to use the pre-extracted salient object region features to represent images. 
    And fine-grained region-level image features and word-level text features are constructed and aligned within the modalities,  respectively. 
    For instance, DVSA in \cite{karpathy2015deep} first adopted R-CNN to detect salient objects and inferred latent alignments between word-level textual features in sentences and region-level visual features in images. 
    Furthermore, to take full advantages of high-level objects and words semantic information, many recent methods \cite{nam2017dual,wu2019learning,ge2021structured,SGRAF,VSRN++} exploited the relationships between the objects and words to help the global embedding of images and sentences, respectively. 
    For instance, \cite{li2019visual,VSRN++} proposed to incorporate the semantic relationship information into visual and textual features by performing object or word relationship reasoning by GCNs. 
    
    \noindent \textbf{Cross-modal interaction retrieval.} 
    Another popular retrieval schemes exploit the fine-grained cross-modal interactions \cite{nam2017dual,lee2018stacked,ji2019saliency,CAMERA,zhang2020context,IMRAM,DIME} to improve the visual-textual semantic alignments.  
    For instance, \cite{DIME} proposed a method for modeling complex dynamic modal interactions based on a three-layer structure with four basic cells per layer. 
    Recently, some works \cite{li2019visual,yi2022multi,GSMN,wang2020cross,Gradual} employed GCNs to improve the interaction and integrate different item representations by a learned graph. 
    Liu \textit{et al.} \cite{GSMN} proposed to learn correspondence of objects and relations between modalities by two different visual and textual reasoning graphs, which is difficult to unify the two modal structures for precise pairing. 
    Long \textit{et al.} \cite{Gradual} also proposed two-modal graphs to help the interactions between modalities, however, the post-interaction concatenation did not substantially improve interactions and additionally introduced word label noise from the scene graph.
    And some works \cite{wang2020cross,nguyen2021deep,Gradual} also encoded the word labels from the detected visual scene graphs causing ambiguity, due to the effect of cross-domain training.


\section{Approach}
    Figure \ref{fig:fig_overall} shows the overall pipeline of our proposed CMSEI for image-sentence retrieval. In this section, we will describe the detailed structure of CMSEI.

    \subsection{Multi-modal Feature Representations} \label{MFX}
        
        \noindent \textbf{Visual representations.}
        To better represent the salient objects and attributes in images, we take advantage of bottom-up-attention model \cite{anderson2018bottom} to extract top-K saliently sub-region features $\hat{I}= \{I_j\}$, $I_j \in \mathbb{R}^{2048}$, based on the category confidence score in an image. 
        Afterward, a fully connected (FC) layer with the parameter $W^o \in \mathbb{R}^{2048 \times D_v}$ is used to project these feature vectors into a $D_v$-dimensional space. 
        Finally, these projected object region features ${V}= \{v_1, \cdots, v_K\}$, $v_j \in \mathbb{R}^{D_v}$, are taken as initial visual representations without semantic enhancement.

        \noindent \textbf{Textual representations.}
        For sentence texts, we follow the recent trends in the community of Natural Language Processing and utilize pre-trained BERT \cite{BERT} model to extract word-level textual representations.
        Similar to visual features processing, we also utilize FC layers to project the extracted word features into a $D_t$-dimensional space, denoted as $T = [t_1, t_2, \cdots, t_N]$, $t_j \in \mathbb{R}^{D_t}$, with length $N$.  
        
        To facilitate cross-modal interaction and embedding space consistency, the projected dimensions are same ($D_v$=$D_t$) for visual and textual representations. 
        For subsequent local-global (image-word/sentence-object) inter-modal interaction and final cross-modal similarity calculation, we use average-pooling operation to obtain the global image feature $\bar{V}$ for sentence-to-image and the global sentence feature $\bar{T}$ for image-to-sentence. 
        
        \subsection{Intra-modal Relationship Enhancement} \label{MSE}
        \noindent \textbf{Explicit visual spatial graph.}
        Since features from the top-K candidate object regions are used for representing the image information, this leads to some regions with semantic overlap but with minor positional bias. 
        In addition, study \cite{cheng2022cross} indicated that the regions with larger Intersection over Union (IoU) as potentially more closely. 
        To this end, following \cite{cheng2022cross}, we also construct explicit spatial non-fully connected graph $G^s=(V,E^s)$ for each image. 
        The semantic similarities and spatial IoUs between sub-regions are combined to represent the adjacency matrix $A^s \in \mathbb{R}^{K \times K}$ as edges for spatial graphs. 
        In particular, if the $IoU_{ij}$ of the $i$-th region and the $j$-th region exceeds the threshold $\mu$, the semantic similarity between them is treated as a weighted edge $A^s_{ij}$, otherwise it is $0$. The pairwise semantic similarity is updated and calculated between regions as: $Sims$ = $(W^v_\varphi V)^T (W^v_\phi V)$ ($W^v_\varphi$ and $W^V_\phi$ denote the mapping parameters).
        For simplicity, we do not explicitly represent the bias term in our paper.

        \noindent \textbf{Explicit visual semantic relationship graph.}
        Different from existing approaches \cite{li2019visual,cheng2022cross} based on implicit relationship graph reasoning, scene graphs have well-defined object relationships, which can overcome the disadvantage of fusing redundant information.
        Unlike approaches \cite{wang2020cross,Gradual} based on scene-graph enhancement, we do not encode the word labels
        predicted by the pre-trained visual scene-graph generator, like \cite{zellers2018neural}. 
        We consider word labels from visual scene graphs of external models have errors and semantically different from the words in the corresponding sentences. 
        This tends to introduce noise that corrupts the cross-modal semantic alignment. 
        In this paper, we construct a non-fully connected semantic relationship graph $G^v=(V^s,E^v)$ between the spatially enhanced objects of each image based on the explicit relationships of the visual scene graphs. 
        In each relationship graph $G^v$, the nodes indicate the object features $V^s$ updated from spatial graph $G^s$ and the edges indicate the existence of semantic associations, as in Figure \ref{fig:frams} (b).
        Here, we construct an adjacency matrix $A^v \in \mathbb{R}^{K \times K} $ to represent these edges for each image, where $A^v_{ij}$=$1$ means $i$-th object is associated with $j$-th object in the semantic relations extracted by a pre-trained visual scene-graph generator and $0$ otherwise.   
        Unlike the spatial graph, since the regions in the scene graph are already strongly correlated, we no longer exploit their semantic similarity.
        
        \noindent \textbf{Visual Feature Embedding.} 
        The currently popular Graph Convolutional Networks (GCNS) \cite{li2019visual} with residuals are used to obtain the final object region features $V^f$, enhanced by updating and embedding of spatial and semantic relationship graphs, named relationship-aware GCNs (R-GCNs), as shown in Figure \ref{fig:fig_overall} \ding{172}. Formally,
        \vspace{-0.5em}
        \begin{equation}
        \begin{array}{c}
            V^s = (A^s V W^s_g) W_{r_1} + V, 
        \end{array}
        \end{equation}
        \vspace{-1.5em}
        \begin{equation}
        \begin{array}{c}
            V^f = ((A^v V W^v_g) W_{r_2} + V^s)W_{r_3} + V,
        \end{array}
        \end{equation}
        where $W^*_g \in \mathbb{R}^{D_v \times D_v}$ are the weight matrix of the GCN layer, $W_{r_*}$ are the residual weights. 
        
        \noindent \textbf{Implicit textual graph building and embedding.} 
        In contrast to the approaches \cite{wang2020cross,nguyen2021deep,cheng2022cross,Gradual} of explicitly modeling inter-word dependencies, we construct a fully connected graph for each sentence, where the semantic features $T$ of the words serve as nodes and the semantic similarities $A^t$ between words serve as edges. 
        We argue that explicit modeling of sentences tends to focus only on the words of object and relation and loses the benefit of many attribute descriptions. 
        Similar to the visual enhancement process, as shown in Figure \ref{fig:fig_overall} \ding{173}, we apply GCNs \cite{li2019visual,VSRN++} with residuals to reason and get the final textual representations $T^f$ with the relationship enhanced, as follows:
        \vspace{-0.5em}
        \begin{equation}
            A^t = (W^t_\varphi T)^T(W^t_\phi T), 
        \end{equation}
        \vspace{-1.5em}
        \begin{equation}
            T^f = (A^t T W^t_g) W_{r_t} + T,
        \end{equation}
        where $W^t_\varphi$ and $W^t_\phi$ denote the mapping parameters, $W_{r_t}$ is the residual weights, $W^t_g$ is the weight matrix of the GCN layer.
        
    \subsection{Inter-modal Interactions} \label{CMI}    
        After image objects and sentence words are reinforced with semantic relationships within a modality, we apply two mainstream inter-modal interaction mechanisms to further enhance the feature representation of the target modality with attention-ware information from another modality. 
        
        \noindent \textbf{Local-local inter-modal interaction.} 
        Similar to literature \cite{lee2018stacked, DIME}, we mine attentions between image objects and sentence words to narrow the semantic gap between two modalities. As shown in Figure \ref{fig:fig_overall} \ding{174}, taking the image-to-sentence example (Due to space limitations and a clearer presentation), we first calculate the cosine similarities for all object-word pairs and calculate the attention weights by a  per-dimension $\lambda$-smoothed Softmax function \cite{chorowski2015attention}, as follows: 
        \vspace{-1em}
        \begin{equation}
           c_{ij} = \frac{({v^f_i})^T t^f_j}{||v^f_i||\ ||t^f_j||}, i \in [1, K], j \in [1,N],
        \end{equation}
        \vspace{-1em}
        \begin{equation}
           \alpha_{ij} = \frac{exp(\lambda c_{ij})}{\sum_{j=1}^N exp(\lambda c_{ij})},  
        \end{equation}
        Finally, we obtain the attended object representation $v_i^t$ $\in$ $V^t$ via a conditional fusion strategy \cite{DIME} from correspondence attention-aware textual vector $q^t_i$ ($q^t_i$=$ \sum_{j=1}^N \alpha_{ij} t^f$), as follows,
        \begin{equation}
            \begin{aligned}
                v_i^t = {\rm ReLU}(W^t_1(v_i^f \odot {\rm Tanh} (W^t_2 q^t_i) + W^t_3 q_i^t)) + v_i^f, \\
            \end{aligned}
        \end{equation}
        where $W^t_*$ are the mapping parameters, ${\rm ReLU}$ and ${\rm Tanh}$ are activation functions. 
        To fully explore fine-grained cross-modal interactions, we perform the above process twice.

        \noindent \textbf{Local-global inter-modal interaction.} 
        As shown in Figure \ref{fig:fig_overall} \ding{175}, we further discover the salience of the fragments in one modality guided by the global contextual information of the other modality, which makes each fragment contains more contextual features. 
        Specifically, for image-to-sentence, we first calculate the semantic similarity between the objects of image $V^t=\{v_1^t, \cdots, v_K^t\}$ and global textual feature $\bar{T}$. 
        Then, we can obtain the relative importance of each object via a sigmoid function. 
        Finally, we add residual connections between the attention-aware object features and the enhanced object features $V^t$, as well as the original features $V$.  
        The above process can be formulated as: 
        \vspace{-0.5em}
        \begin{equation}
           r_{i} = {\rm \sigma}(W^r v_i^t \odot \bar{T}),
        \end{equation}
        \vspace{-1.5em}
        \begin{equation}
          v^o_i = r_{i}v_i^t + v_i^t + {\rm ReLU} (v_i) ,
        \end{equation}
	    where $W^r$ denotes the mapping parameter. Similarly, for sentence-to-image, we enhance the word features via calculating the relative importance of each word between the words of the sentence and the global image feature $\bar{V}$. 
	    
	    To obtain the final match score between image and sentence, we average and normalise the final object features of the image and calculate the cosine similarity with the global sentence features.

    \subsection{Objective Function} \label{OBJ}
        In the above training process, all the parameters can be simultaneously optimized by minimizing a bidirectional triplet ranking loss \cite{faghri2017vse++}, when aligning the image and sentence as follows:
        \begin{equation}
            \begin{aligned}
                \mathcal{L}_{rank}(I,{T}) = \sum_{(I,\hat{T})} [\nabla - {\rm cos}(I,T)+ {\rm cos}(I,\hat{T})]_{+} \\
                + \sum_{(\hat{I},T)} [\nabla - {\rm cos}(I,T)+ {\rm cos}(\hat{I},T)]_{+} 
            \end{aligned}
        \end{equation}     
        where $\nabla$ serves as a margin constraint, ${\rm cos}(\cdot,\cdot)$ indicates cosine similarity function, and $[\cdot]_+ = {\rm max}(0, \cdot)$. Note that, $(I,S)$ denotes the given matched image-sentence pair and its corresponding negative samples are denoted as $\hat{I}$ and $\hat{S}$, respectively.

\begin{table*}[t]
\scriptsize
\begin{center}
\fontsize{9}{11.5}\selectfont
\renewcommand\tabcolsep{10.0pt}
\caption{Comparisons of experimental results on MS-COCO 5-folds 1K test set and full 5K test set.} \label{tab:tab1_coco}
\begin{tabular}{c|cccccc|c}
\hline
\hline
\multicolumn{1}{c|}{\multirow{2}{*}{Method}} & \multicolumn{3}{c}{Sentence Retrieval} & \multicolumn{3}{c|}{Image Retrieval}      & \multirow{2}{*}{rSum} \\
\multicolumn{1}{c|}{}                        & R@1        & R@5       & R@10      & R@1  & R@5 & \multicolumn{1}{c|}{R@10} &                        \\ \hline
\multicolumn{8}{c}{5-folds 1K}                                                                                                                              \\ \hline

    SCAN$\rm ^*_{ECCV'18}$ \cite{lee2018stacked}        & 72.7    &  94.8     &  98.4     &  58.8     &  88.4    &  \multicolumn{1}{c|}{94.8}  & 507.9       \\ 
    VSRN$\rm ^*_{ICCV'19}$ \cite{li2019visual}    & 76.2   & 94.8  & 98.2   & 62.8 & 89.7 &{95.1}     & 516.8  \\ 

    IMRAM$\rm ^*_{CVPR'20}$ \cite{IMRAM}      & 76.7    & 95.6    & 98.5    & 61.7    &   89.1   & 95.0  & 516.6     \\ 
    CAAN$\rm _{CVPR'20}$ \cite{zhang2020context}       & 75.5    &   95.4     &  98.5    &  61.3    &   89.7    & {95.2}  & 515.6     \\ 
    GSMN$\rm ^*_{CVPR'20}$ \cite{GSMN} & 78.4    & 96.4   & 98.6     & 63.3     & 90.1     &  95.7      & 522.5  \\
    
    CAMERA$\rm ^*_{ACMMM'20}$ \cite{CAMERA} & 78.0    & 95.1   & 97.9     & 60.3     & 85.9     & 91.7      & 508.9  \\
    SGRAF$\rm ^*_{AAAI'21}$ \cite{SGRAF}  & 79.6    & 96.2    & 98.5     & 63.2     & 90.7     & 96.1     & 524.3  \\
    VSE$\rm \infty_{CVPR'21}$ \cite{VSEwuqiong} & 79.7    & 96.4    & \textbf{98.9}     & 64.8     & 91.4     & 96.3     & \underline{527.5}  \\
    DIME$\rm ^*_{SIGIR'21}$ \cite{DIME}   & 78.8     & 96.3    & 98.7     & 64.8     & 91.5     & 96.5     & 526.6 \\ 
    CGMN$\rm ^*_{TOMM'22}$ \cite{cheng2022cross}   & 76.8 & 95.4 & 98.3 & 63.8 & 90.7 & 95.7 & 520.7  \\ 
    VSRN++$\rm ^*_{TPAMI'22}$ \cite{VSRN++}  & 77.9    & 96.0     & 98.5   &  64.1     & 91.0   & 96.1  & 523.6  \\ 
    GraDual$\rm ^*_{WACV'22}$ \cite{Gradual} & 77.0    & 96.4   & 98.6     & \underline{65.3}     & \textbf{91.9}     & 96.4      & 525.6  \\ 
    NAAF$\rm ^*_{CVPR'22}$ \cite{NAAF} & \underline{80.5}    & \underline{96.5}      & 98.8     & 64.1     & 90.7     & \underline{96.5}      & 527.2  \\ 
  \hline
  
\multicolumn{1}{c|}{CMSEI$^*$ (ours)}  & \textbf{81.4}    & \textbf{96.6}      & \underline{98.8}     & \textbf{65.8}     & \underline{91.8}     & \textbf{96.8}      & \textbf{531.1}  \\  \hline

\hline

\multicolumn{8}{c}{Full 5K}                                                                                                                              \\ \hline
\multicolumn{1}{c|}{VSE++$\rm _{BMVC'18}$ \cite{faghri2017vse++}} & 41.3   & 69.2  & 81.2   & 30.3 & 59.1 & \multicolumn{1}{c|}{72.4}     & 353.5  \\
\multicolumn{1}{c|}{SCAN$\rm ^*_{ECCV'18}$ \cite{lee2018stacked}}  & 50.4   & 82.2  & 90.0   & 38.6 & 69.3 & \multicolumn{1}{c|}{80.4}  & 410.9  \\ 

    VSRN$\rm ^*_{ICCV'19}$ \cite{li2019visual}    & 53.0  & 81.1   & 89.4  & 40.5    & 70.6   & 81.1     & 415.7  \\ 
    IMRAM$\rm ^*_{CVPR'20}$ \cite{IMRAM}   & 53.7 &  83.2  &  91.0 & 39.7    & 69.1   & 79.8     & 416.5    \\ 
    CAAN$\rm _{CVPR'020}$ \cite{zhang2020context}   & 52.5 &  83.3    & 90.9   & 41.2  &  70.3   & 82.9  &  421.1   \\ 
    CAMERA$\rm ^*_{ACMMM'20}$ \cite{CAMERA}  & 55.1    & 82.9    & 91.2     & 40.5     & 71.7     & 82.5    & 423.9  \\
    VSE$\rm \infty_{CVPR'21}$ \cite{VSEwuqiong}  & 58.3   & 85.3 & \underline{92.3}    & 42.4     &  72.7   &  \underline{83.2}     & 434.3  \\
    DIME$\rm _{SIGIR'21}$ \cite{DIME}    & \underline{59.3}   & \underline{85.4}   & 91.9    & \underline{43.1}   &  \underline{73.0}   &  83.1     & \underline{435.8}\\ 
    CGMN$\rm ^*_{TOMM'22}$ \cite{cheng2022cross}  &  53.4  &  81.3  &  89.6  &  41.2  &  71.9  &  82.4  &  419.8  \\ 
    VSRN++$\rm ^*_{TPAMI'22}$ \cite{VSRN++}   & 54.7 & 82.9   & 90.9    & 42.0   & 72.2    & 82.7     & 425.4  \\ 
    NAAF$\rm ^*_{CVPR'22}$ \cite{NAAF}    & 58.9   & 85.2   & 92.0   & 42.5   & 70.9   & 81.4   &  430.9  \\ 
\hline
\multicolumn{1}{c|}{CMSEI$^*$ (ours)}  & \textbf{61.5}    & \textbf{86.3}      & \textbf{92.7}     & \textbf{44.0}     & \textbf{73.4}     & \textbf{83.4}      & \textbf{441.2}  \\ 
\hline
\hline
\end{tabular}
\end{center}
\vspace{-3em}
\end{table*}

\section{Experiments}
    In this section, we report the results of our experiments to evaluate the proposed approach, CMSEI. 
    We will introduce the dataset and experimental settings first. 
    Then, CMSEI is compared with state-of-the-art image-sentence retrieval approaches quantitatively. 
    Finally, we qualitatively analyze the results in detail. 
    \subsection{Dataset and Evaluation Metrics}
        \noindent \textbf{Dataset.}
        We evaluate our proposed approach on the MS-COCO \cite{lin2014microsoft} and Flickr30k \cite{young2014f30k} datasets, which are the most popular benchmark datasets for image-sentence retrieval task.
       There are over 123,000 images in MS-COCO. 
        Following the splits of most existing methods \cite{ADAPT,Gradual,VSRN++,VSEwuqiong,DIME}, there are 113,287 images for training, 5,000 images for validation and 5000 images for testing. 
        On MS-COCO, we report results on both 5-folder 1K and full 5K test sets, which are the average results of 5 folds of 1K test images and the results of full 5K test set, respectively.
        Flickr30K contains over 31,000 images with 29, 000 images for the training, 1,000 images for the testing, and 1,014 images for the validation. 
        Each image in these two benchmarks is given five corresponding sentences by different AMT workers. 
        
        \noindent\textbf{Evaluation metrics.}
        Following the standard evaluation protocol, we employ the widely-used recall metric, R@K (K=1,5,10) evaluation metric, which denotes the percentage of ground-truth being matched at top K results, respectively.
        Moreover, we report the ``\textit{rSum}'' criterion that sums up all six recall rates of R@K, which provides a more comprehensive evaluation to testify the overall performance. 

    \subsection{Implementation Details}
        Our model is trained on a single TITAN RTX GPU with 24 GB memory.
        The whole network except the Faster-RCNN model \cite{ren2015faster} is trained from scratch with the default initializer of PyTorch using ADAM optimizer \cite{kingma2014adam} a mini-batch size 64. 
        The learning rate is set to 0.0002 initially with a decay rate of 0.1 every 15 epochs. Maximum epoch number is set to 30. 
        The margin of triplet ranking loss $\nabla$ is set to 0.2. 
        The threshold $\mu$ is set to 0.4. 
        For the visual object features, Top-K (K=36) object regions are selected with the highest class detection confidence scores. 
        The visual scene graphs are generated by Neural Motifs \cite{zellers2018neural}, and we use the maximum IoU to find the corresponding regions in the original Top-K salient regions. 
        The initial dimensions of visual and textual embedding space are set to 2048 and 768 respectively, which are transformed to the same 1024-dimensional (\textit{i.e.}, $D_v$= $D_s$=1024). 
        The most dimensions of mapping parameters are set to 1024-dimensional.

    \subsection{Comparison with State-of-the-art Methods}
        \noindent \textbf{Baseline and state-of-the-arts.} We compare our proposed CMSEI with several image-sentence retrieval methods on the MS-COCO and Flickr30K  datasets in Table \ref{tab:tab1_coco} and Table \ref{tab:tab1_F30k}, including (1) the global matching methods, \textit{i.e.}, VSE++ \cite{faghri2017vse++}, SGRAF \cite{SGRAF}, VSE$\infty$ \cite{VSEwuqiong} (the reported version with same object inputs), and (2) the attention-based cross-modal interaction methods, \textit{i.e.}, SCAN$^*$ \cite{lee2018stacked}, CAAN \cite{zhang2020context}, IMRAM$^*$ \cite{IMRAM}, DIME \cite{DIME} \textit{etc.}, and (3) the graph-based retrieval methods, \textit{i.e.}, VSRN \cite{li2019visual}, CGMN \cite{cheng2022cross} and GraDual \cite{Gradual}, and (4) latest state-of-the-art methods, \textit{i.e.}, DIME \cite{DIME}, NAAF\cite{NAAF}, \textit{etc.} 
        Note that, the ensemble models with ``*" are further improved due to the complementarity between multiple models. For fair comparison, we also provide the ensemble results, which are averaged similarity scores of image-text model and text-image model. And the results of each single model are provided in Table \ref{tab:Single}. 

\begin{table*}[t]
\scriptsize
\vspace{-0.2em}
\begin{center}
\fontsize{9}{11.5}\selectfont
\renewcommand\tabcolsep{10.0pt}
\caption{Comparisons of experimental results on Flickr30K 1K test set. '$^*$' indicates the performance of an ensemble model. } \label{tab:tab1_F30k}
\begin{tabular}{c|cccccc|c}
\hline
\hline
\multicolumn{1}{c|}{\multirow{2}{*}{Method}} & \multicolumn{3}{c}{Sentence Retrieval} & \multicolumn{3}{c|}{Image Retrieval}      & \multirow{2}{*}{rSum} \\
\multicolumn{1}{c|}{}                        & R@1        & R@5       & R@10      & R@1  & R@5 & \multicolumn{1}{c|}{R@10} &                        \\ \hline
    SCAN$\rm ^*_{ECCV'18}$ \cite{lee2018stacked}        & 67.4          & 90.3          & 95.8          & 48.6          & 77.7          & 85.2          & 465.0                 \\ 
     VSRN$\rm ^*_{ICCV'19}$ \cite{li2019visual}    & 71.3   & 90.6  & 96.0   & 54.7 &81.8 & {88.2}     & 482.6  \\
    CAAN$\rm _{CVPR'20}$ \cite{zhang2020context}       & 70.1          & 91.6          & 97.2  & 52.8          & 79.0          & 87.9          & 478.6                 \\ 
    IMRAM$\rm ^*_{CVPR'20}$ \cite{IMRAM} & 74.1    & 93.0   & 96.6     & 53.9     & 79.4     &  87.2      & 484.2  \\
    GSMN$\rm ^*_{CVPR'20}$ \cite{GSMN} & 76.4    & 94.3   & 97.3     & 57.4     & 82.3     &  89.0      & 496.8  \\
    CAMERA$\rm ^*_{ACMMM'20}$ \cite{CAMERA}  & 78.0    & 95.1   & 97.9     & 60.3     & 85.9     & 91.7      & 508.9  \\
    SHAN $\rm ^*_{IJCAI'21}$ \cite{SHAN} & 74.6   & 93.5   & 96.9     & 55.3     & 81.3     & 88.4      & 490.0  \\
    SGRAF$\rm ^*_{AAAI'21}$ \cite{SGRAF} & 77.8    & 94.1   & 97.4     & 58.5     & 83.0     & 88.8      & 499.6  \\
    VSE$\rm \infty_{CVPR'21}$ \cite{VSEwuqiong} & 81.7    & 95.4   & 97.6     & 61.4     & 85.9     & 91.5      & 513.5  \\
    DIME$\rm ^*_{SIGIR'21}$ \cite{DIME} & 81.0    & \underline{95.9}   & \underline{98.4}     & 63.6     & \textbf{88.1}     & \textbf{93.0}      &  \underline{520.0}  \\ 
    CGMN$\rm ^*_{TOMM'22}$ \cite{cheng2022cross}   & 77.9 &  93.8 & 96.8 & 59.9 & 85.1 & 90.6 & 504.1  \\ 
    
    VSRN++$\rm ^*_{TPAMI'22}$ \cite{VSRN++} & 79.2    & 94.6   & 97.5     & 60.6     & 85.6     & 91.4      & 508.9  \\ 
    GraDual$\rm ^*_{WACV'22}$ \cite{Gradual} & 78.3    & 96.0   & 98.0     & \underline{64.0}     & 86.7     & 92.0      & 511.4  \\ 
    NAAF$\rm ^*_{CVPR'22}$ \cite{NAAF} & \underline{81.9}    & 96.1   & 98.3     & 61.0     & 85.3     & 90.6      & 513.2  \\ 
    
    \hline
    CMSEI$^*$ (ours)      & \textbf{82.3}  & \textbf{96.4} & \textbf{98.6}  & \textbf{64.1} & \underline{87.3} & \underline{92.6} & \textbf{521.3}  \\ \hline  
\hline
\end{tabular}
\end{center}
\vspace{-2.5em}
\end{table*}

\begin{table}[t]
\scriptsize
\begin{center}
\fontsize{9}{11.5}\selectfont
\renewcommand\tabcolsep{4.0pt}
\caption{The results of our single CMSEI model, image-to-sentence (I-T) and sentence-to-image (T-I) versions, on MS-COCO and Flickr30K.} \label{tab:Single}
\begin{tabular}{c|cccccc}
\hline
\hline
\multicolumn{1}{c|}{\multirow{2}{*}{Version}} & \multicolumn{3}{c}{Sentence Retrieval} & \multicolumn{3}{c}{Image Retrieval}   \\
\multicolumn{1}{c|}{}                        & R@1        & R@5       & R@10      & R@1  & R@5 & \multicolumn{1}{c}{R@10}  \\ \hline
\multicolumn{7}{c}{MS-COCO 5-folds 1K}    \\ \hline 
    I-T      &  78.7  &  95.9   &  98.3   &   62.5 &   90.5 &   96.1\\
    T-I      &  78.7 &  96.2 &  98.8      &  63.6  &  90.9  &  96.2           \\
    \hline
    \multicolumn{7}{c}{Flickr30K}    \\ \hline 
    I-T      &  78.0 &  95.2 &  98.0  &  59.5  &  84.2  &  91.1 \\
    T-I       &  79.0  &  94.6  &  97.9  &   60.4 &  85.5 &  90.5           \\
    \hline
\hline
\end{tabular}
\end{center}
\vspace{-2.7em}
\end{table}

        \noindent \textbf{Quantitative comparison on MS-COCO.}
        Table \ref{tab:tab1_coco} lists the experimental results on two kinds of MS-COCO test sets, 5-folds 1K (at the top) and full 5K (at the bottom).
        Specifically, compared with the state-of-the-art model NAAF \cite{NAAF} on MS-COCO 1K test set, our CMSEI achieves 0.9\% and 1.7\% improvements in terms of R@1 on both image and sentence retrieval, respectively. 
        Compared with the best cross-modal interaction method DIME \cite{DIME}, CMSEI achieves 2.6\% and 1.0\% improvements in terms of R@1 on image and sentence retrieval, respectively. 
        And CMSEI clearly outperforms the methods GraDual \cite{Gradual} and CGMN \cite{cheng2022cross}, which also employ graph networks, by 5.5\% and 10.4\% in terms of \textit{rSum}, respectively.
        
        Furthermore, on the larger image-sentence retrieval test data (MS-COCO Full 5K test set), including 5000 images and 25000 sentences, CMSEI outperforms recent methods with a large gap. 
        Following the common protocol \cite{DIME,NAAF},  CMSEI achieves 5.3\%, 15.8\%, and 10.3\% improvements in terms of \textit{rSum} compared with the latest state-of-the-arts DIME \cite{DIME}, VSRN++ \cite{VSRN++} and NAAF \cite{NAAF}, respectively. 
        It clearly demonstrates the powerful effectiveness of the proposed CMSEI model with the huge improvements. 
        
        \noindent \textbf{Quantitative comparison on Flickr30K.} 
        Quantitative results on Flickr30K 1K test set are shown in Table \ref{tab:tab1_F30k}, where the proposed approach CMSEI outperforms all state-of-the-art methods in terms of \textit{rSum}. 
        Though for individual recall metrics, we see variations in performance, however, the proposed CMSEI shows clear improvements under all most metrics compared with the latest state-of-the-art methods.  

        \noindent \textbf{Generalization ability for domain adaptation.} 
        We further validate the generalization ability of the proposed CMSEI on challenging cross-datasets, which is meaningful for evaluating the cross-modal retrieval performance in real-scenario. 
        Specifically, similar to CVSE \cite{wang2020consensus}, we transfer our model trained on MS-COCO to Flickr30K dataset. 
        As shown in Table \ref{tab:tab_transfor}, the proposed CMSEI achieves significantly outperforms the baselines. 
        It reflects that CMSEI has an excellent capability of generalization for cross-dataset image-sentence retrieval. 

\begin{table}[t]
\scriptsize
\begin{center}
\fontsize{9}{11.5}\selectfont
\renewcommand\tabcolsep{5.0pt}
\caption{Comparison results on cross-dataset generalization from MS-COCO to Flickr30k. ${\dag}$ means the results are obtained from their published pre-trained model.} \label{tab:tab_transfor}
\begin{tabular}{c|cccc}
\hline
\hline
\multicolumn{1}{c|}{\multirow{2}{*}{Method}} & \multicolumn{2}{c}{Sentence Retrieval} & \multicolumn{2}{c}{Image Retrieval}   \\
\multicolumn{1}{c|}{}                        & R@1       & R@10      & R@1  & \multicolumn{1}{c}{R@10}  \\ \hline

    VSE++$\rm _{BMVC'18}$ \cite{faghri2017vse++}       & 40.5	& 77.7	& 28.4	& 66.6 \\
    SCAN$\rm ^*_{ECCV2018}$ \cite{lee2018stacked}	    & 49.8	& 86.0	& 38.4	& 74.4 \\
    CVSE$\rm _{ECCV'20}$ \cite{wang2020consensus}	    & 57.8	& 87.2	& 44.8	& 81.1 \\
    VSE$\rm \infty^{\dag}_{CVPR'21}$ \cite{VSEwuqiong}   & \underline{68.0}	& 93.7	& 50.0	& 84.9 \\
    DIME$\rm ^{*\dag}_{SIGIR'21}$ \cite{DIME}  & 67.4	& \underline{94.5}	& \underline{53.7}	& \underline{86.5} \\
    \hline 	
    CMSEI (ours)       & \textbf{69.6}  & \textbf{95.2}  & \textbf{53.7} & \textbf{87.2} \\ \hline
\hline
\end{tabular}
\end{center}
\vspace{-3em}
\end{table}    
\begin{figure}[t] 
	\centering
	\vspace{-0.5em}
	\includegraphics[width=1.0\linewidth]{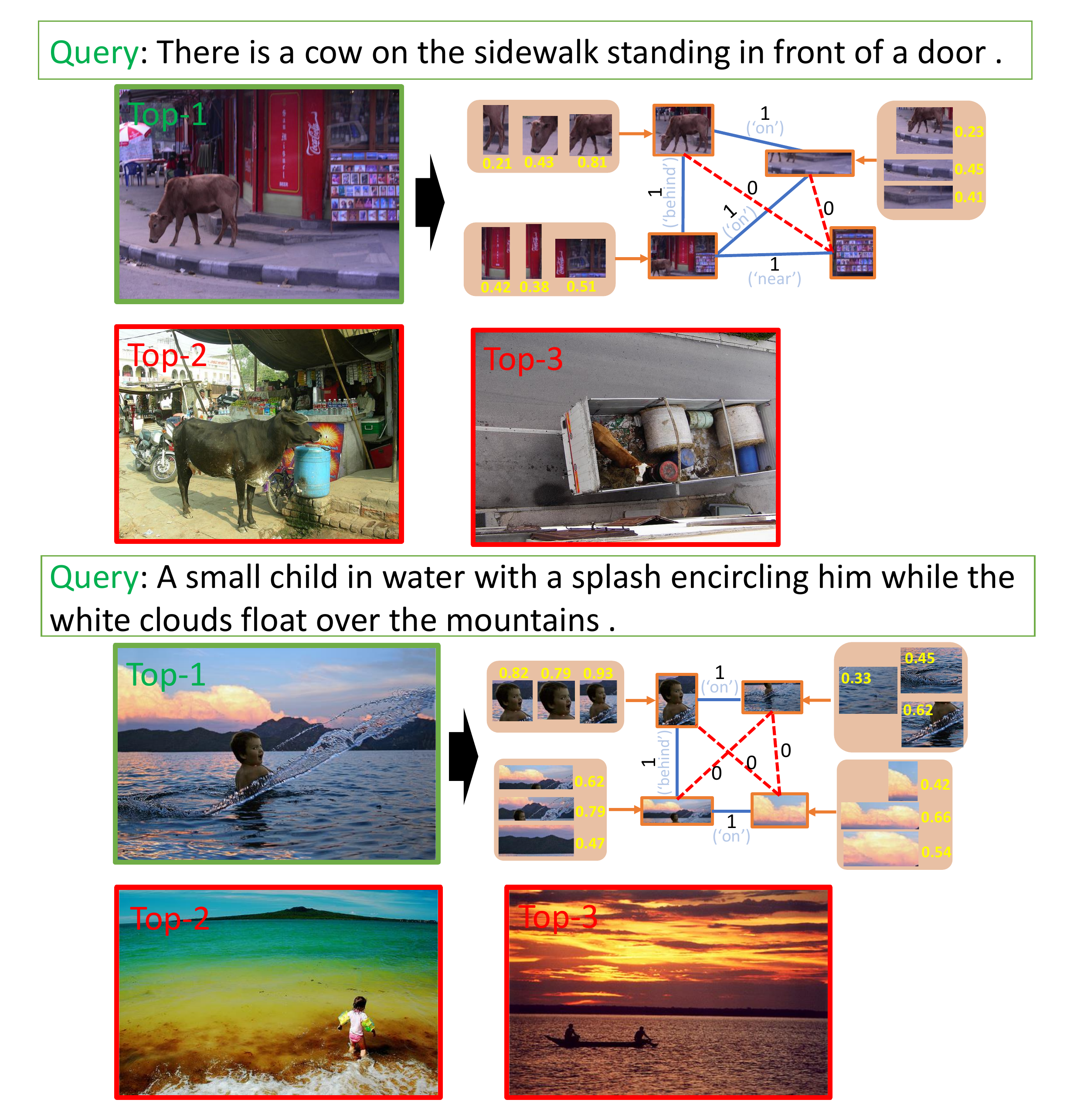}
	\vspace{-1.5em}
	\caption{visualization of Top-3 image retrieval results of our CMSEI on MS-COCO (at the top) and Flickr30K (at the bottom).  The correctly matched images are marked in green and the mismatched images are marked in red.  The learned similarities between objects with high spacial IoUs and the explicit semantic relationship graphs (blue lines mean semantic correlations (indicated by 1), red dashed lines are no significant semantic correlations (indicated by 0)) for the matched image fragments are also partially presented (best viewed in color).}
	\label{fig:example_t2i}
	\vspace{-1.5em}
\end{figure}

\begin{figure}[t] 
	\centering
	\vspace{-1em}
	\includegraphics[width=1\linewidth]{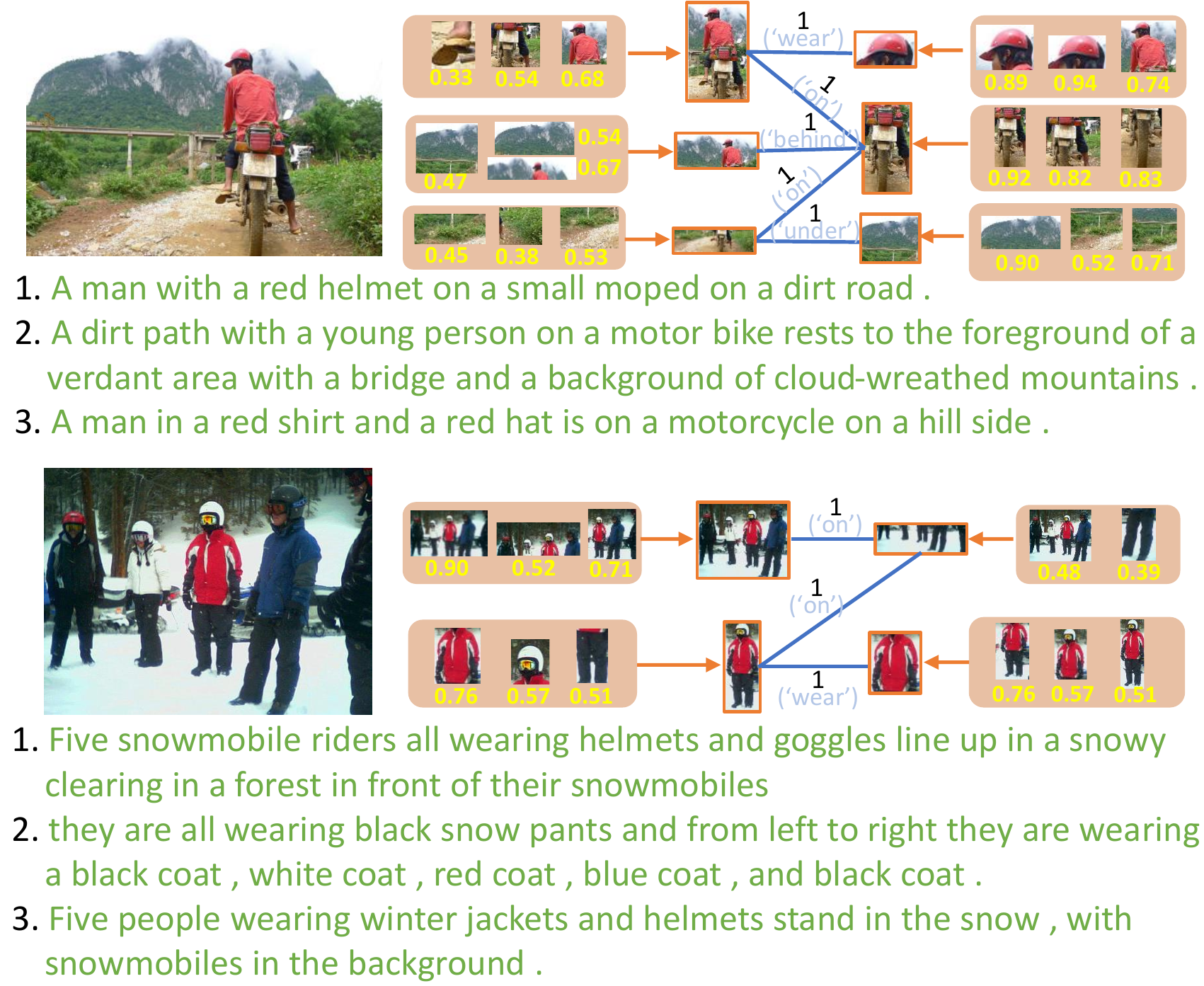}
	\vspace{-1.3em}
	\caption{visualization of Top-3 sentence retrieval results of our CMSEI on MS-COCO (at the top) and Flickr30K (at the bottom). The corresponding explicit relationship graphs with the relevant and correlation weights among fragments of images are also partially presented (best viewed in color).} 
	\label{fig:example_i2t}
	\vspace{-1.2em}
\end{figure}

        \noindent \textbf{Visualization of results.} 
        To better understand the effectiveness of the proposed CMSEI, we visualize matching results from image and sentence retrieval on both MS-COCO and Flickr30K in Figure \ref{fig:example_t2i} and Figure \ref{fig:example_i2t}, respectively. 
        The example on top is from MS-COCO and the one below is from Flick30k. 
        Moreover, we visualize the explicit visual spacial and semantic relationship graphs (due to limitations of space we show the graphs partially) for the corresponding images for both retrieval directions, which are used in CMSEI. 
        We also show the learned similarities between objects with high spacial IoUs in space.
        And the structured correlations among the objects (explicit correlation weight is 1, otherwise 0) can be maintained with the explicit semantic correlation graph guidance. 
        It can be observed that the proposed relationship graphs provide more precise spatial and semantic correlations between the object regions, which can help the model to interact more comprehensively. 

    \subsection{Ablation Studies}
    We perform detailed ablation studies on Flickr30K to investigate the effectiveness of each component of our proposed CMSEI. 
    
    \noindent \textbf{Effects of visual spatial graph.} 
    In Table \ref{tab:tab3_ab}, CMSEI decreases absolutely by 2.3\% on Flickr30K in terms of $rSum$ when removing the visual spatial graph (w/o VSG). 
    It suggests that the spatial graph reasoning plays an important role in concentrating on spatially relevant regional features for fragments in images. 
    In addition, we achieved slightly lower results using self-attention networks (w. SA) \cite{vaswani2017attention}, an implicit relationship modeling method, as an alternative to VSG. 
    It demonstrates that our proposed visual spatial graph reasoning can effectively aggregate spatially relevant regional features compared to implicit relational reasoning based on self-attention.

\begin{table}[t]
\scriptsize
\vspace{-0.5em}
\begin{center}
\fontsize{9}{12}\selectfont
\renewcommand\tabcolsep{2.0pt}
\caption{Ablation studies on Flickr30K 1K test set. All values are ensemble results by averaging two models' (I-T and T-I) similarity.} \label{tab:tab3_ab}
\begin{tabular}{c|cccccc|c}
\hline
\hline
\multicolumn{1}{c|}{\multirow{2}{*}{Method}} & \multicolumn{3}{c}{Sentence Retrieval} & \multicolumn{3}{c|}{Image Retrieval}  & \multirow{2}{*}{rSum} \\
\multicolumn{1}{c|}{}                        & R@1        & R@5       & R@10      & R@1  & R@5 & \multicolumn{1}{c|}{R@10}  \\ \hline
    w/o VSG          &  81.5   &  95.8    & 98.6    & 63.5    & 87.1    & 92.5    & 519.0       \\
    w. SA (VSG)      &  81.2   &  96.2    & 98.3    & 63.5    & 87.2     & 92.6   & 519.0 \\
    w/o VSRG         &  79.5   &  95.2    &  98.0   &  62.3   &  87.0    &  92.3  &514.8       \\
    w. FG (VSRG)     &  81.0   & 95.4     & 98.6    & 63.0    & 87.3     &  92.5  &517.8   \\
    w/o TG           &  79.7   &  95.0    &  97.9   & 61.8    & 86.7     &  92.2  &513.3  \\
    w. DTG (TG)      &  79.6   & 95.7     & 97.7    & 61.6    &  86.9    & 92.4   &513.9  \\
    w/o LLII         &  73.5   &  93.6    & 96.7    &  57.5   &  84.2    &  90.5    & 496.2       \\
    w/o LGII         &  80.0   &  95.2    &  98.2  & 62.9     &  87.2    &  92.3  & 515.8   \\
    \hline
    CMSEI       & \textbf{82.3}  & \textbf{96.4} & \textbf{98.6}  & \textbf{64.1} & \textbf{87.3} & \textbf{92.6}  & \textbf{521.3} \\ \hline
\hline
\end{tabular}
\end{center}
\vspace{-2.5em}
\end{table}

    \noindent \textbf{Effects of explicit visual semantic relationship graph.} 
    As shown in Table \ref{tab:tab3_ab}, CMSEI decreases absolutely 3.5\% in terms of $rSum$ on Flickr30k when replacing explicit visual semantic relationship graph (VSRG) by a fully-connected graph (indicated by w. FG) for images. 
    When dropping VSRG directly (indicated by w/o VSRG), it degrades the $rSum$ by a clear  7.0\%. 
    These observations suggest that our explicit VSRG effectively improve visual semantic feature embedding and avoid irrelevant feature incorporation.

    \noindent \textbf{Effects of implicit textual graph.} 
    When dropping the textual graph reasoning (w/o TG) of sentences, a significant drop in results can be observed. 
    Without implicit semantic reasoning on a fully connected textual graph, we construct a semantic dependency text graph (indicated by w. DTG) by the same way as in method \cite{cheng2022cross}, resulting in great degradation. 
    We speculate that these dependencies lose some textual information when interacting across modalities.

    \noindent \textbf{Effects of local-local and local-global inter-modal interactions.} 
    We evaluate the impact of the local-local and local-global inter-modal interaction (LLII and LGII) for CMSEI. 
    As shown in Table \ref{tab:tab3_ab}, the absence of LLII and the absence of LGII reduce 4.2\% and 1.0\% in terms of the average of all metrics on Flickr30k, respectively. 
    It is obvious that the multiple inter-modal interactions play a vital role in image-sentence retrieval process, which also suggests that cross-modal interactions effectively narrow the semantic gap between the two modalities.

\section{Conclusion}
In this paper, we present a cross-modal semantic enhanced interaction method (CMSEI) for image-sentence retrieval. 
CMSEI engages in (i) enhancing the visual semantic representation with the inter-object relationships and (ii) enhancing the visual and textual semantic representation with multi-level joint semantic correlations on intra-fragment, inter-fragment, and inter-instance. 
To this end, we propose the intra- and inter-modal semantic correlations and optimize the integrated structured model with cross-modal semantic alignment in an end-to-end representation learning way.
Extensive quantitative comparisons demonstrate that our CMSEI achieves state-of-the-art performance on the most of standard evaluation metrics across MS-COCO and Flickr30K benchmarks.

{\small
\bibliographystyle{ieee_fullname}
\bibliography{egbib}
}

\end{document}